\documentclass[conference]{IEEEtran}
\IEEEoverridecommandlockouts
\usepackage{cite}
\usepackage{amsmath,amssymb,amsfonts}
\usepackage{algorithmic}
\usepackage{graphicx}
\usepackage{textcomp}
\usepackage{xcolor}
\def\BibTeX{{\rm B\kern-.05em{\sc i\kern-.025em b}\kern-.08em
    T\kern-.1667em\lower.7ex\hbox{E}\kern-.125emX}}
\begin{document}

\title{Attention-based Bidirectional LSTM for Deceptive Opinion Spam Classification}
% {\footnotesize \textsuperscript{*}Note: Sub-titles are not captured in Xplore and
% should not be used}
% \thanks{Identify applicable funding agency here. If none, delete this.}
% }

\author{\IEEEauthorblockN{Ashish Salunkhe}
\IEEEauthorblockA{\textit{Persistent Systems} \\
% \textit{name of organization (of Aff.)}\\
Pune, India \\
avsalunkhe98@gmail.com}
% \and
% \IEEEauthorblockN{2\textsuperscript{nd} Given Name Surname}
% \IEEEauthorblockA{\textit{dept. name of organization (of Aff.)} \\
% \textit{name of organization (of Aff.)}\\
% City, Country \\
% email address or ORCID}
% \and
% \IEEEauthorblockN{3\textsuperscript{rd} Given Name Surname}
% \IEEEauthorblockA{\textit{dept. name of organization (of Aff.)} \\
% \textit{name of organization (of Aff.)}\\
% City, Country \\
% email address or ORCID}
% \and
% \IEEEauthorblockN{4\textsuperscript{th} Given Name Surname}
% \IEEEauthorblockA{\textit{dept. name of organization (of Aff.)} \\
% \textit{name of organization (of Aff.)}\\
% City, Country \\
% email address or ORCID}
% \and
% \IEEEauthorblockN{5\textsuperscript{th} Given Name Surname}
% \IEEEauthorblockA{\textit{dept. name of organization (of Aff.)} \\
% \textit{name of organization (of Aff.)}\\
% City, Country \\
% email address or ORCID}
% \and
% \IEEEauthorblockN{6\textsuperscript{th} Given Name Surname}
% \IEEEauthorblockA{\textit{dept. name of organization (of Aff.)} \\
% \textit{name of organization (of Aff.)}\\
% City, Country \\
% email address or ORCID}
}

\maketitle

\begin{abstract}
Online Reviews play a vital role in e-commerce for decision-making.  Much of the population makes the decision of which places, restaurant to visit, what to buy and from where to buy based on the reviews posted on the respective platforms. A fraudulent review or opinion spam is categorized as an untruthful or deceptive review. Positive reviews of a product or a restaurant helps attract customers and thereby lead to an increase in sales whereas negative reviews may hamper the progress of a restaurant or sales of a product and thereby lead to defamed reputation and loss.
Fraudulent reviews are deliberately posted on various online review platforms to trick customers to buy, visit or distract against a product or a restaurant.  They are also written to commend or discredit the product’s repute. 
The work aims at detecting and classifying the reviews as fraudulent/deceptive or truthful. It involves use of various deep learning techniques for classifying the reviews and an overview of proposed approach involving Attention-based Bidirectional LSTM to tackle issues related to semantic information in reviews and a comparative study over baseline machine learning techniques for review classification.
\end{abstract}


\section{Introduction}
With the advent of the internet, a large depot for digital information and user-generated content is available. Such content is generated through community interactions, papers, documents, opinions expressed on various platforms, social media websites and many other sources. Content like reviews, helps businesses improve their sales and grow. With an increase in reviews on various review platforms, customers rely on more on online reviews for decision making. Also, clients use online reviews for decision making in context to their future business partners. Thus, online reviews may help businesses grow or harm their reputations. Opinion spams are such online reviews that either harm or help businesses grow by posting fake reviews on the online review platforms. Detecting opinion spams or fraudulent reviews manually is slog altogether.
Human beings are not always able to classify truthful and fake reviews. Current research has questioned the credibility of online reviews. For instance, around 20\% of reviews posted on Yelp are found to be faked by paid human writers \cite{yao2017automated}. Opinion spamming has opened easy monetization.
Spam reviews can be classified into two types: \cite{Jindal08opinionspam} Misleading reviews and non reviews. Misleading reviews are either undeserving positive reviews or unjust negative reviews regarding a product or a service. Non reviews are those which contain no opinions about the product, rather, they either promote irrelevant advertisements or comments. 

There has been ample research in developing review spam classification systems. Past researches in text classification make use of labeled data for text classification. Awad and Elseuofi applied various machine learning techniques for the automated classification of emails \cite{awad2011machine}. Manlangit et al. proposed an intelligent system which could detect fraud transactions using classification algorithms \cite{manlangit2017efficient}. One of the major issues faced for developing such classifier systems for review platforms is the imbalance in data. The frequency of spam samples can be sparse in comparison to non-spam samples. Conventional Machine learning algorithms when trained on such data can have a bias towards majority class which will lead to a new sample being accepted as relevant. This ambiguity can be resolved using balanced dataset. This can be achieved using either undersampling the majority class or oversampling the minority class.

In existing research, comparison of results is limited due to consideration of different features, models or data. In this paper, an approach is proposed that involves deceptive reviews which are classified using a supervised learning approach on a balanced data set. The classification is treated as a text classification problem. The reviews are classified into two classes - Deceptive and Truthful. The work makes use of labeled data to plot a decision boundary based on which a new test case can be classified into the appropriate class. Text preprocessing techniques and different models are employed to successfully detect review spams. This includes deep-learning models that recently produced state-of-the-art benchmarks on text classification problems. Along with this, word embeddings are used to train continuous word representations. These pre-trained word vectors improve the performance of the deep learning models, but results show a linear classifier based on frequency-based embeddings outperforms the other approaches along with the deep learning models.

The remainder of the paper is structured as follows: Section 2 provides an overview of related work. Along with the overview of Cleaning and pre-processing techniques used prior to classification and the proposed approach is described in Section 3. Description of the data set, evaluation metrics and experimental results are provided in Section 4. Finally, Section 5 draws the conclusion and suggests future direction of the work.
\section{Related Work}
 
Fake  Review Detection has involved various approaches based on review content, user-reviewer behavior, and sentiment analysis in classifying user reviews. Review classification problem was first put forth by Jindal and Liu \cite{Jindal08opinionspam}. They classified reviews into three categories:\cite{Jindal08opinionspam} \newline 
Type 1: False Opinions: undeserving positive reviews or unjust
negative reviews. \newline
Type  2:  Reviews  on  Brand:  Reviews  on  the  brand  of  the
product instead of the product itself. \newline
Type  3:  Non-Reviews:  They  are texts  with  random content like an advertisement which is irrelevant to the product. \newline
Jindal and Liu have suggested that outlier reviews must be identified as suspicious reviews \cite{Jindal08opinionspam}.  Yoo  and  Gretzel  (2009) gathered 40 truthful and 42 deceptive hotel reviews and manually compared the linguistic differences between them. \cite{10.1007/978-3-211-93971-0_4}. Ott et. al (2011) employed Turkers to write fake reviews and created a benchmark data set \cite{ott2011finding}. Their data set was adopted by the further line of work \cite{Ott:2012:EPD:2187836.2187864} Feng et al. (2012) looked into syntactic features from context-free grammar parse trees \cite{feng2012syntactic} to improve the classification performance. Recently, deep neural networks have shown promising results in text classification problems.  They have shown good results in learning underlying features
in literature.  Most  characteristic  progress  has  been  made  by Zeng et Al.(2014), who used Convolutional Neural Networks for relation classification\cite{zeng2014relation}.  Since  CNN doesn’t consider the context of the sequence and also it considers words only in close range,  another approach is suggested by Mikolov et  Al.(2010) based on Recurrent  Neural  Networks\cite{mikolov2010recurrent}.  Another related work proposed by Zhang and Wang  (2015)  involved use of Bidirectional  RNN\cite{zhang2015relation}.  But  Bidirectional  RNN faces a vanishing gradient problem.  Hence,  to  resolve  this  problem,
Long  short  term  Memory  was  introduced  by  Hochreiter  and Schmidhuber (1997)\cite{hochreiter1997long}.

Finally, proposed work is based on and related to three approaches
-  one  architecture  involving  Bidirectional  Long  short  term
memory proposed by Zhang et al. (2015)\cite{zhang2015relation}, another involving
Hierarchical  Attention  network  proposed  by  Zichao  Yang
et  al. (2016)\cite{Yang2016HierarchicalAN}  and  finally  Attention-Based  Bidirectional  Long Short-Term Memory Networks for Relation Classification proposed by Peng Zhou et al. (2016) \cite{zhou2016attention}.
\section{Proposed Approach}
Classification of reviews can be done via two categories: 
Review centric features and Reviewer centric features.
The proposed approach focuses on review centric features. 
Classification of reviews is done using neural network model which learns continuous vector representations for documents of variable lengths, which is used as features to classify a spam or not for each document. Also, comparative study of methods of text
classification for opinion spam. 
\subsection{Preprocessing}
Data obtained from online review platforms cannot be fed directly into the machine learning algorithm.
To extract useful information from the data, different preprocessing techniques need to be applied on the data set. The textual data in Deceptive Opinion Spam Corpus\cite{Ott:2012:EPD:2187836.2187864}\cite{ott2011finding} is converted into feature vectors which is passed to the classification algorithms.
Various Preprocessing techniques include:
  \begin{itemize}
    \item Encoding the classes (Deceptive = 1 and Truthful = 0)
    \item Removing numeric and empty texts
    \item Removing punctuation from texts
    \item Convert words to lower case
    \item Stop word Removal: Words like are, is, they, this etc can be found in both the classes (Deceptive and Truthful) and thus they cannot be a deciding factor for classification. Such words are known as stop words and we need to remove them form the text.
    \item Stemming: Reduce inflectional forms of a word to a common root form.\cite{jivani2011comparative} \item Tokenization: The process of breaking down the character sequence or a defined document unit into individual entities, called tokens, and simultaneously removing punctuation.
    \item Vectorization: The filtered text is converted into a sparse matrix of tokens by vectorizing them. Common vectorizing techniques include Count Vectorizer and Tf-Idf vectorizer.
    The Tf-Idf weight is composed by two terms: Term Frequency and Inverse Document Frequency. The term frequency(TF) gives the number of times a word appears in a sample, divided by the total number of words in that sample. The Inverse Document Frequency (IDF), computed as the logarithm of the number of the samples in the data set divided by the number of samples where the specific word appears.
    Mathematical Equation for Tf-Idf follows:
    \begin{equation}
    tfidf_{i,d} = \frac{n_{i,j}}{\sum_k n_{k,j}} \cdot \log{\frac{|D|}{|{d : t_i \in d}|}}
    \end{equation}
    \end{itemize}
The above mentioned preprocessing techniques are implemented
using the nltk, beautiful soup and scikit learn packages in
Python 3.6. After carrying out these techniques, the text data
is converted to a sparse matrix consisting of numbers
which can be fed to a classification algorithm. As a part of
the evaluation process,the data is split into train and test
sets with the model being trained on 80\% of the data and
results are calculated for the remaining 20\% of data set. For
any new test case, the above preprocessing methods still need
to be carried out and the class label is predicted for the feature
vector of the test case.
\subsection{Word Representations}
Word representations are either fixed or distributed. In this paper, distributed word representations are used. Distributed word representations represent word as features vectors. They show semantical or syntactical dependencies. For instance, 'car' and 'automobile' can have similar vector values and low numerical difference since they have related meanings. Hence, useful properties can be obtained using such linear relationship.
Word representations used in the work:
\begin{itemize}
    \item Word2Vec: Word2Vec model is composed of shallow neural network models called CBOW (Continuous Bag of Words) and Skip-Gram; used to learn word embeddings\cite{mikolov2013distributed}.
    This model constructs vocabulary from training data set and then prepares word embedding vectors. Pre-trained vectors can be loaded published by Google trained on part of Google News Data set. The model consists off 300-dimensional vectors for 3-million words and phrases.
    \item GloVe: GloVe is an unsupervised algorithm for obtaining word embedding vectors\cite{pennington2014glove}. Training is performed on grouped global word-word co-occurrence statistics from a corpus, and the resulting representations showcase linear substructures of the word vector space. \cite{pennington2014glove}
    \newline GloVe has been used to generate word vector.
    
\end{itemize}
\subsection{Model Building}
The next step is to train the classifier using the features created in previous step.
Different machine learning models can be used to train the final model. Different classifiers are implemented for this purpose and compare their results with performance metrics and choose the final model for the classification.\newline
Following are the classifiers implemented:
\begin{itemize}
\item Naive Bayes Classifier: MultinomialNB
\item Linear Classifier: Logistic Regression
\item Support Vector Machine
\item Stochastic Gradient Descent 
\item Deep Neural Networks
    \begin{itemize}
        \item Convolutional Neural Network (CNN)
\item Long Short Term Model (LSTM)
\item Bidirectional LSTM
\item Recurrent CNN
\item Bidirectional LSTM Attention
    \end{itemize}
\end{itemize}
\subsubsection{Naive Bayes Classifier}
Naive Bayes illustrates a supervised learning technique. It is a classification technique based on Bayes' Theorem. It assumes independence among predictors. In layman's terms it assumes presence of particular feature in class is unrelated to the other features' presence. Naive Bayes model is easy to build and performs well on large data sets. It is known to outperform sophisticated classification models.
\subsubsection{Linear Classifier - Logistic Regression}
Logistic regression is a linear classifier which measures the relationship between the categorical dependent variable and one or more independent variables. It estimates the probabilities using a logistic function. 
\subsubsection{Support Vector Machine} Support Vector Machine exemplifies supervised learning technique used for both classification and regression which finds the hyperplane that best segregates two classes. 
\subsubsection{Stochastic Gradient Descent} Stochastic Gradient Descent is a efficient approach to discriminative learning of linear classifiers like SVM and Logistic Regression under convex loss functions. SGD has been successfully used on large data sets and it's easy implementation and efficiency has helped solve sparse machine learning problems in text classification and natural language processing. 

 Naive Bayes Classifier, Logistic Regression, Support Vector Machine, Stochastic Gradient Descent are trained on features such as count vectors, word level TF-IDF vectors, N-gram level TF-IDF vectors and Character level TF-IDF vectors and calculate the performance metrics for further comparative study and abalative analysis. 
 
 \subsubsection{Deep Neural Networks}
 Deep Learning models have been successful in natural language processing tasks like  distributed word learning, sentence and document representation \cite{mikolov2013efficient}, parsing \cite{socher2013parsing}, statistical machine translation \cite{devlin2012statistical}, sentiment classification \cite{agarwal2016sentiment}\cite{kim2014convolutional}, opinion-spam classification \cite{Ott:2012:EPD:2187836.2187864}\cite{ott2013negative}\cite{ott2011finding}. 
 
\begin{center}
\centering
    \begin{table*}[t]
\def\arraystretch{1.5}
\centering
            \begin{tabular}{|c | c | c | c | c | c |}                          % little tabular example 
            \hline
               Sr. No. & Model & Accuracy (\%) & Precision Score & Recall Score & F1 Score\\
                \hline
                1 & MultinomialNB & 0.9025 & 0.9325 & 0.8601 & 0.8948 \\
                \hline
                 2 & Stochastic Gradient Descent (SGD) &  0.8775    &   0.8913    &   0.8497    & 0.8700\\
                \hline
                3 & Logistic Regression &    0.8700   &  0.8691   &   0.8601    & 0.8645\\
                \hline
                 4 & Support Vector Machine  &  0.5625     &   0.525    &   0.9792    & 0.6835\\
                \hline

            \end{tabular}
            \newline
\caption{Performance of various classification algorithms on Deceptive Opinion Spam Corpus}
\end{table*}
\end{center}

In this paper, the focus is on opinion spam classification in online reviews using deep learning based approaches. 
Proposed method is compared with widely used text classification
methods and the state-of-the-art approaches is done.
The approach presents six different architectures to achieve this. Then exploratory analysis is performed and a comparative study over previous architectures and architectures like Attention-CNN, Attention Bidirectional LSTM and a hybrid of both approaches as Bidirectional LSTM-CNN-Attention architecture is done. These architectures have shown promising results in the other text-classification tasks \cite{Abreu_2019,adhikari2019docbert,chorowski2015attention,agarwal2016sentiment,kim2014convolutional,markov2016author}. Thus, we try to implement these architectures alongside the proposed architecture.
\begin{itemize}
    \item \textbf{Convolutional Neural Network:}
Text classification using embeddings and Convolutional Neural Networks was previously introduced by Yoon Kim \cite{kim2014convolutional}. Similary, CNN is used along with GloVe 100-Dimension embeddings. The convolutional kernel concatenates the embeddings into predefined window to compute the output. Subsequent layers apply filters and combine the output for the classification.  Pre-trained vectors can be loaded published by Google trained on part of Google News Data set. The model consists off 300-dimensional vectors for 3-million words and phrases.
    \item \textbf{Recurrent Neural Networks - LSTM and  Bidirectional LSTM:}  TextCNN works well for text classification since it considers words in close range. But it doesn't consider the context of the text sequence. It doesn't learn the sequential structure of the text sequence where words are depended on the previous words in a sentence. Recurrent Neural Networks are specialized for this purpose. They remember information for an extended period using the hidden states and connect it to the current task. While LSTM works in one direction, Bidirectional LSTM retains the contextual information in both directions. Firstly, GloVe 100-Dimension embedding layer is used with the LSTM neural network. Secondly, Bidirectional LSTM is used along with GloVe 50-Dimension embeddings.
    \item \textbf{Recurrent CNN - CNN-Bidirectional LSTM:} A hybrid architecture consisting of CNN and Bidirectional LSTM and Doc2Vec and Tf-Idf vectors as embeddings instead of GloVe embeddings is used to obtain neural-network-based document embeddings as introduced in \cite{markov2016author}. It gives a fixed-length vector as output.
    \item \textbf{Attention-based Bidirectional LSTM:} We propose Attention-based Bidirectional LSTM for the mentioned problem. Important information can appear at any position in the text sequence. To capture the most important semantic information in the text sequence Attention-Bidirectional LSTM is used as introduced in \cite{zhang2015relation}. \newline 
Model can be summarized as:
\begin{enumerate}
    \item Input layer: Input Sequence
    \item Embedding layer
    \item BiLSTM layer
    \item Attention layer
    \item Output layer
\end{enumerate}
\item \textbf{Hierarchical Attention:} Attention neural networks have recently shown promising results in text classification, machine translation, question answering and other application areas. \cite{bahdanau2014neural} \cite{chorowski2015attention}\cite{hermann2015teaching}. Attention mechanism is used for text classification task as proposed by Peng Zhou, Wei Shi, et al. in \cite{zhou2016attention}. Attention model consists of two parts as shown in figures 1 and 2: Bidirectional LSTM and Attention Networks. Bidirectional Recurrent Neural Networks are stacked on word level followed by attention networks to extract only the important word representations required to get the meaning of the sentence and later aggregate these important word representations to form sentence vectors. Similarly, only important sentence vectors are aggregated to get the meaning of the document and generate a document vector which will be further passed for text classification.  
    
\end{itemize}
\begin{figure}[h]
    \centering
    \includegraphics[width=\linewidth]{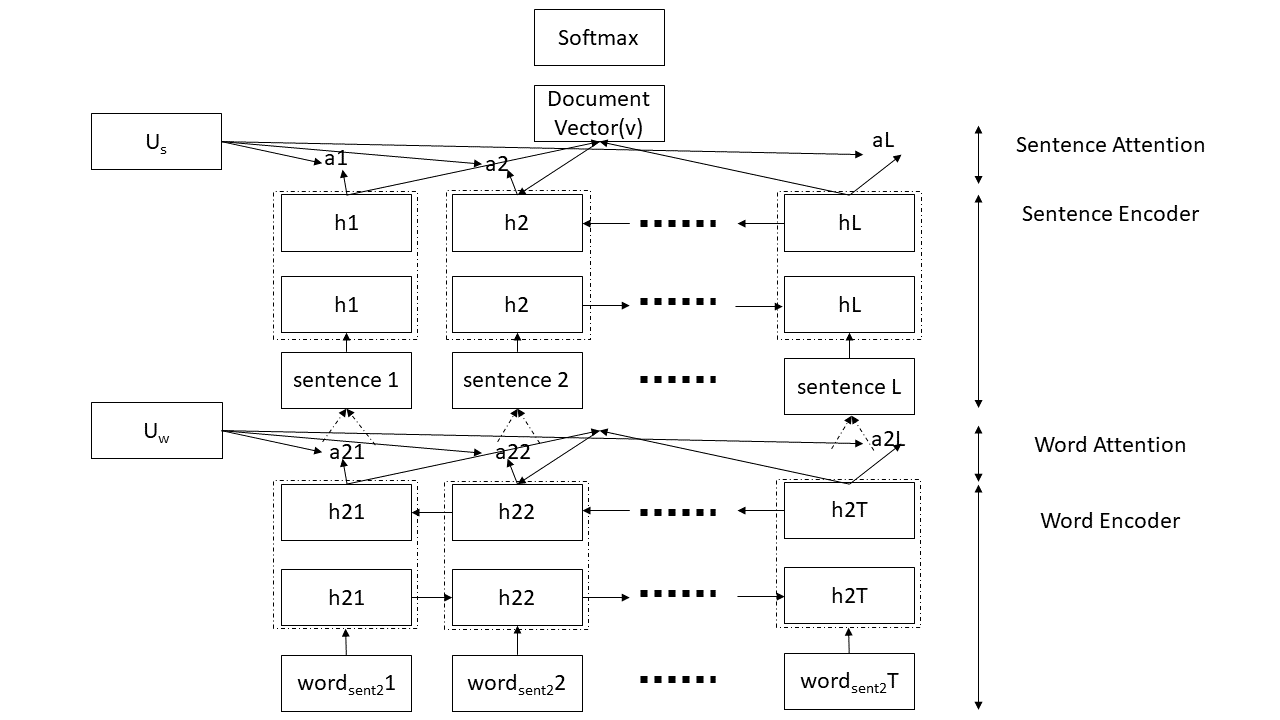}
    \caption{Hierarchical Attention Model. Adapted from \cite{Yang2016HierarchicalAN}}
    \label{fig:han}
\end{figure}
\begin{figure}[h]
    \centering
    \includegraphics[width=\linewidth]{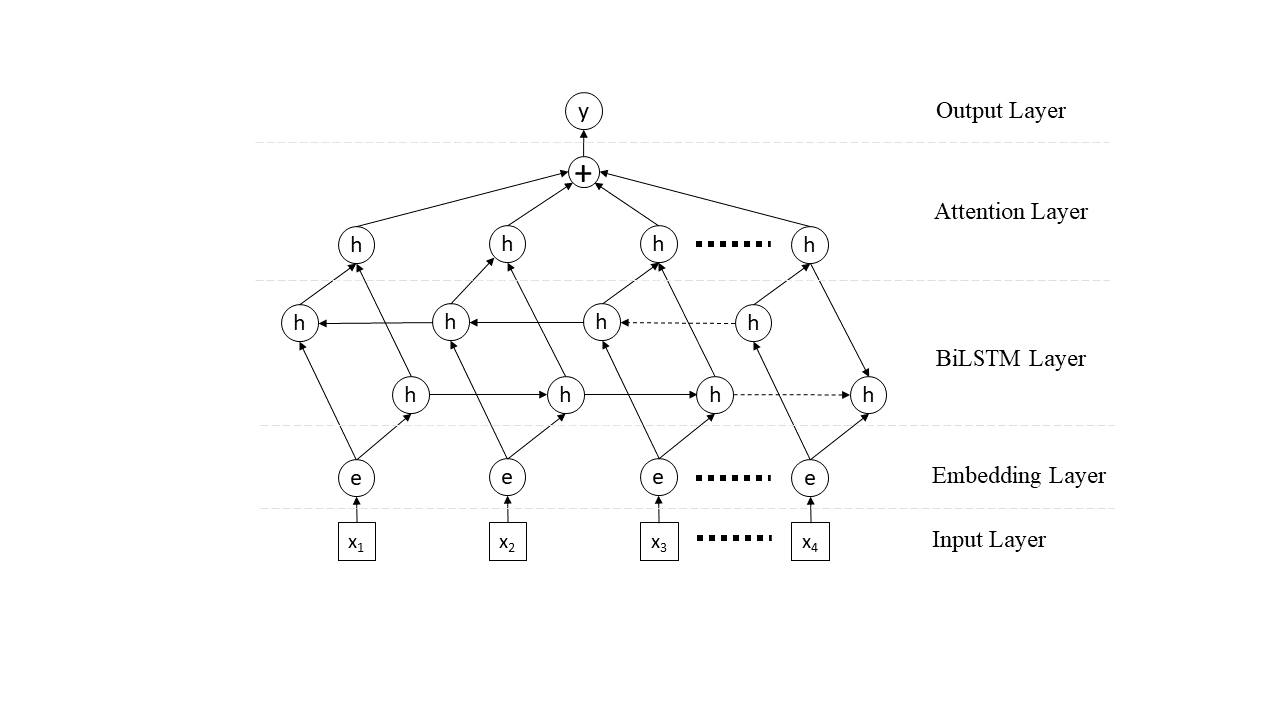}
    \caption{Attention-based BiLSTM. Adapted from \cite{zhou2016attention}}
    \label{fig:han}
\end{figure}

\section{Evaluation}

\subsection{Dataset Description}
To study and compare results of various approaches benchmark dataset Deceptive Opinion Spam Corpus consisting of 1600 hotel reviews created by Ott et. al \cite{Ott:2012:EPD:2187836.2187864}\cite{ott2013negative}\cite{ott2011finding} was used.\newline
This corpus contains:
\begin{itemize}
\item 400 truthful positive reviews from TripAdvisor (described in \cite{ott2011finding})
\item 400 deceptive positive reviews from Mechanical Turk (described in \cite{ott2011finding})
\item 400 truthful negative reviews from Expedia, Hotels.com, Orbitz, Priceline, TripAdvisor and Yelp (described in \cite{ott2013negative})
\item 400 deceptive negative reviews from Mechanical Turk (described in \cite{ott2013negative})
\end{itemize}

Each of the above datasets consist of 20 reviews for each of the 20 most popular Chicago hotels \cite{ott2011finding}.
The dataset consists of features as hotel, polarity, source, text and target value as deceptive. 

\subsection{Performance Metrics}
The classification by linear models is evaluated by the measure of accuracy, precision, recall and F1 Score. 
\begin{center}
    Accuracy = $\frac{\displaystyle TP + TN}{\displaystyle TP + TN + FP + FN}$\newline\newline
    Precision = $\frac{\displaystyle TP}{\displaystyle TP + FP}$ \newline\newline
    Recall = $\frac{\displaystyle TP}{\displaystyle TP + FN}$
    \newline\newline
    F1 Score = 2 * $\frac{\displaystyle Precision * Recall}{\displaystyle Precision + Recall}$
\end{center}
where,  TP: True positive(outcome where the model correctly predicts the positive class),
TN: True negative(outcome where the model correctly predicts the negative class),
FP: False positive(outcome where the model incorrectly predicts the positive class),
FN: False negative(outcome where the model incorrectly predicts the negative class)

\begin{table}[h]
    \centering
    \begin{tabular}{|l|c|c|}
\hline
& Deceptive & Truthful\\
\hline
Deceptive & TP & FP\\
\hline
Truthful & FN & TN\\
\hline

\end{tabular}
            \newline
    \label{tab:my_label}
\end{table}
%---------------------------------------------------------------------------------------------
\begin{table}[h]
    \centering
    \begin{tabular}{|c | c | c | c |}                          % little tabular example 
            \hline
                Model & Train accuracy & Test accuracy \\ [0.5ex]
               \hline % inserts single horizontal line
 Bidirectional LSTM + GLoVe(50D) & 92.17 & 88.13 \\ [0.5ex] \hline % inserting body of the table 
 LSTM + GLoVe(100D) & 99.18 & 85.75 \\ [0.5ex] \hline
 CNN + LSTM + Doc2Vec +TF-IDF & 96.23 & 92.19 \\ [0.5ex] \hline
 BiLSTM + Attention + GLoVe(100D) & 99.18 & 90.25 \\ [0.5ex] \hline
            \end{tabular}
            \newline
    \caption{Performance of Neural Networks on Deceptive Opinion Spam Corpus}
    \label{tab:my_label}
\end{table}
%---------------------------------------------------------------------------------------------
\subsection{Ablative analysis}
The use of review-centric features like n-gram TF-IDF vectors and character level TF-IDF vectors can achieve satisfactory results
%---------------------------------------------------------------------------------------------
\begin{table}[h]
    \centering
     \begin{tabular}{| c | c | c | c |}                          % little tabular example 
            \hline
                Model & Accuracy (\%) & AUC Score & F1 Score\\
                \hline
                MNB + N-Gram   & 0.845   &  0.918 &  0.8393\\
                \hline
                 MNB + CharLevel  & 0.8025 &   0.914    & 0.7893  \\ 
                 \hline
                 LR + N-Gram  & 0.7975 &  0.912   &   0.8480   \\
                \hline
                LR + CharLevel  & 0.8225 & 0.916 & 0.8461 \\
                \hline

            \end{tabular}
            \newline
    \caption{Ablative Analysis through review centric features}
    \label{tab:my_label}
\end{table}

%-------------------------------------------------------
\subsection{Evaluation of Results}
The training dataset is passed through various classification algorithms namely: Multinomial Naive Bayes, Logistic regression, Stochastic Gradient Descent and Support Vector Machine. The results based on metrics like accuracy, precision score, recall and F1 score are presented in Table 1.\newline
SVM: After being trained on textual data, it gives the least F1 score of 68.35\%, least accuracy of 56.25\%, least precision of 52.5\% and highest recall 97.92\% \newline
SGD and MultinomialNB: 
The accuracy when considering text feature is observed to be 90\% for Multinomial Naïve Bayes and around 88\% for Stochastic Gradient Descent. From table 1, it can be clearly concluded that the Multinomial Naïve Bayes classifier performs better than the other classification algorithms.
Logistic Regression too performs better with a accuracy of 87\% and F1 score of 86.45\%

Amongst all deep learning architectures used Recurrent Convolutional Neural Network architecture and Attention-based Bidirectional Long Short Term Memory show better results with validation accuracy of 92.19\% and 90.25\%  respectively. It has better results than most of the existing approaches involving Multinomial Naive Bayes, Logistic regression, Stochastic Gradient Descent and Support Vector Machine and other deep learning architectures. 
%---------------------------------------------------------
\section{Conclusion}
It was observed that classification algorithms achieve satisfactory results on review-centric features like n-gram TF-IDF vectors and character level TF-IDF vectors through the abalative analysis.
Experimental results demonstrate that the model performs significantly better than previous methods using deep neural networks. Visualization of the attention-based approach adapted from \cite{Yang2016HierarchicalAN} demonstrates that the model is effective in capturing important words and sentences. The performance of the approach can be enhanced for all deep learning architectures through hyper-parameter tuning.
%-------------------------------------------------------

\section*{Acknowledgment}

This work was performed during the summer internship at CoreView Systems Pvt. Ltd., Pune, India while being student at Pimpri Chinchwad College of Engineering and Research, Pune.  I would like to extend my special thanks to Mr. Makarand Vaidya, CEO, CoreView Systems Pvt. Ltd. for his guidance, encouragement and useful critiques for this research work.

\end{document}